# QUALITY ASSESSMENT OF PIXEL-LEVEL IMAGE FUSION USING FUZZY LOGIC

Srinivasa Rao Dammavalam[1] Seetha Maddala[2] and Krishna Prasad MHM[3]

[1]Department of Information Technology, VNRVJIET, Hyderabad, AP
dammavalam2@gmail.com

[2]Department of CSE, GNITS, Hyderabad, AP
seetha.maddala@gmail.com

[3] Department of CSE, JNTU College of Engineering, Vizianagaram, India
krishnaprasad.mhm@gmail.com

## ABSTRACT

*Image fusion is to reduce uncertainty and minimize redundancy in the output while maximizing relevant information from two or more images of a scene into a single composite image that is more informative and is more suitable for visual perception or processing tasks like medical imaging, remote sensing, concealed weapon detection, weather forecasting, biometrics etc. Image fusion combines registered images to produce a high quality fused image with spatial and spectral information. The fused image with more information will improve the performance of image analysis algorithms used in different applications. In this paper, we proposed a fuzzy logic method to fuse images from different sensors, in order to enhance the quality and compared proposed method with two other methods i.e. image fusion using wavelet transform and weighted average discrete wavelet transform based image fusion using genetic algorithm (here onwards abbreviated as GA) along with quality evaluation parameters image quality index (IQI), mutual information measure ( MIM), root mean square error (RMSE), peak signal to noise ratio (PSNR), fusion factor (FF), fusion symmetry (FS) and fusion index (FI) and entropy. The results obtained from proposed fuzzy based image fusion approach improves quality of fused image as compared to earlier reported methods, wavelet transform based image fusion and weighted average discrete wavelet transform based image fusion using genetic algorithm.*

## 1. INTRODUCTION

In [1] proposed a method for image fusion where the fused image is obtained by inverse transforming a synthetic wavelet transform array which combines information from the two input images. A medical image fusion based on discrete wavelet transform using Java technology approach described to combine the salient feature of images obtained from different compatible medical devices and integrated this method into a distributed application [2]. In [3] a novel image fusion scheme based on biorthogonal wavelet decomposition is presented in which two images are decomposed into sub-images with different frequency, and information fusion is performed using these images under the certain criterion, and finally these sub-images are reconstructed into the result image with plentiful information. In [4] an introduction to wavelet

 

transform theory and an overview of image fusion techniques are given, and the results from a number of wavelet-based image fusion approaches are compared and it has been proved that, in general, wavelet-based schemes perform better while minimizing color distortion. A novel architecture with a hybrid algorithm is defined in which pixel based maximum selection rule to low frequency approximations and filter mask based fusion to high frequency details of wavelet decomposition is applied [5]. A Region based Pan Sharpening Method using Match Measure and Fuzzy Logic approach provides novel trade off solution to preserve spectral and spatial quality using fuzzy logic in which match measure, region based approach and fuzzy logic methods are combined to produce quality Pan sharp image [6]. In [7] proposed a theoretical framework mimicking the aggregation process, based on the use of fuzzy logic approach, fusion operators to enrich the classical fusion process with the introduction of spatial information modelling. Multisensor image fusion was proposed for surveillance systems in which fuzzy logic modelling utilized to fuse images from different sensors, in order to enhance visualization for surveillance systems [8]. In [9] a novel hybrid multispectral image fusion based on fuzzy logic approach is proposed using combine framework of wavelet transform and fuzzy logic and it provides novel solution between the spectral and spatial fidelity and preserves more detail spectral and spatial information. Fuzzy logic based image fusion for multi-view though-the-wall radar technique proposed where global fusion operator considered and it is desirable to consider the differences between each pixel using a local operator [10]. In [11] a image fusion algorithm based on fuzzy logic and wavelet was proposed and was aimed at the visible and infrared image fusion and addresses an algorithm based on the discrete wavelet transform and fuzzy logic approach. In [12] a new algorithm is proposed for sharpening multi-spectral images using their corresponding high-resolution panchromatic images. It uses a nonlinear fuzzy fusion rule to combine features extracted from the original images.

## 2. IMAGE FUSION USING WAVELET TRANSFORM

Wavelet transform is a type signal representation that can give the frequency content of the signal at a particular instant of time. In [13] the wavelet based image fusion process proposed in which steps mainly involved are registering source images, performing wavelet transform on each input images, then generating a fusion decision map based on a defined fusion rule and constructing fused wavelet coefficient map from the wavelet coefficients of the input images according to the fusion decision map. Finally, transform back to the spatial domain. Image fusion based on wavelet transform is the most commonly used approach, which fuses the source images information in wavelet domain according to some fusion rules. The block diagram of a generic wavelet based image fusion approach is shown in the Fig1 [14].

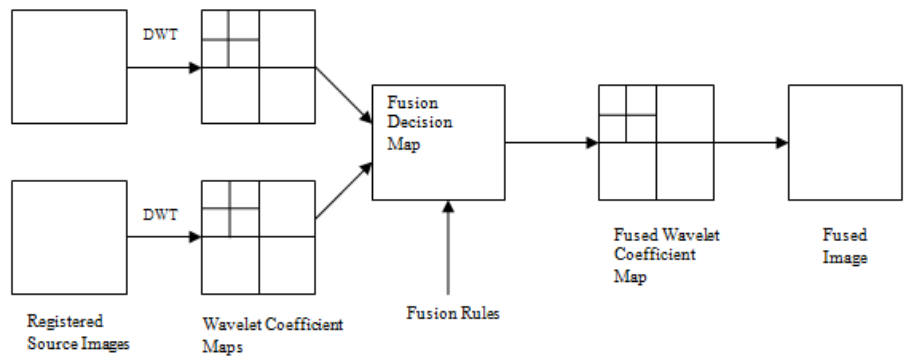

Figure 1. The generic structures of the image fusion using wavelet transform

## 2.1. Wavelet Based Image Fusion Algorithm

Wavelets are localized waves .They have finite energy. They are suited for analyses of transient signal.They are finite duration oscillatory functions with zero average value. The irregularity and good localization properties make them better basis for analysis of signals with discontinuities. The steps involved in wavelet based image fusion algorithm are as following [15]

1. Read the two input images, I1 and I2 to be fused.

2. Perform independent wavelet decomposition of the two images

3. Apply pixel based algorithm for approximations which involves fusion based on taking the maximum valued pixels from approximations of source images I1and I2

4. Based on the maximum valued pixels between the approximations, a binary decision
map is generated gives the decision rule for fusion of approximation coefficients in the two source images I1and I2.

5. The final fused transform corresponding to approximations through maximum selection pixel rule is obtained.

6. Concatenation of fused approximations and details gives the new coefficient matrix.

7. Apply inverse wavelet transform to reconstruct the resultant fused image and display the result.

## 3. IMAGE FUSION USING WEIGHTED AVERAGE DWT USING GA

GA's are being used in different applications such as function Optimization, Image Processing, Parameter Optimization of Controllers, Multi-Objective Optimization, etc. In [16] fusion of panchromatic and multispectral images by genetic algorithm is adopted to define the injection model which establishes how the missing high pass information is extracted from the pan image and the fitness function of the algorithm which provides the algorithm parameters driving the image fusion process is based on a quality index specifically. A region based multi-focus image fusion algorithm using spatial frequency and genetic algorithm was introduced and the basic idea is to divide the source images into blocks, and then select the corresponding blocks with higher spatial frequency value to construct the resultant fused image in which GA is designed for quality assessment of MS images [17].

The procedural steps in Genetic Algorithm are given as follows [18].

• Choose initial population

• Evaluate the fitness of each individual in population

• Repeat

• Select best-ranking individuals to reproduce a new Population

• Breed new generation through crossover and mutation to give birth to offspring

- Evaluate the individual fitness of the offspring

- Replace worst ranked part of population with offspring

- Until some termination condition is met

### 3.1. Image Fusion using Discrete Wavelet Transform (DWT)

Image fusion using wavelet scheme decomposes source images I1(a, b) and I2(a, b) into approximation and detailed coefficients at required level using DWT. The approximation and detailed coefficients of both images are combined using fusion rule. The fused image could be obtained by taking the inverse discrete wavelet transform (IDWT) as:

$$I(a,b) = \frac{DWT\{I1(a,b)\} + DWT\{I2(a,b)\}}{2} \quad (1)$$

The fusion rule used here is simply averages the approximation coefficients and picks the detailed coefficient in each sub band with the largest magnitude.

### 3.2. Image Fusion using Weighted Average DWT

In this approach additional weights are selected along with the DWT of the images. The fused image can be obtained by taking the inverse discrete wavelet transform (IDWT) as:

$$I(a,b) = \frac{W1*DWT\{I1(a,b)\} + W2*DWT\{I2(a,b)\}}{W1+W2} \quad (2)$$

### 3.3. Weighted Average DWT based Image Fusion using GA

In this process additional weights are estimated using GA along with the DWT of the images. The fused image can be obtained by taking the inverse discrete wavelet transform (IDWT) as:

$$I(a,b) = \frac{GA(W1)*DWT\{I1(a,b)\} + GA(W2)*DWT\{I2(a,b)\}}{GA(W1)+GA(W2)} \quad (3)$$

## 4. IMAGE FUSION USING FUZZY LOGIC APPROACH

The set of input images used in this algorithm are registered images. Registration process gives correspondence between images and ensured that spatial correspondence established, fusion makes sense [19].

### 4.1. Fuzzy Logic in Image Processing

Fuzzy image processing is not a unique theory. It is a collection different fuzzy approaches that understand, represent and process the images, their segments and features as fuzzy sets. The representation and processing depend on the selected fuzzy technique and on the problem to be solved. Fuzzy logic has three main stages like image fuzzification, modification of membership values and image defuzzification.

The coding of image data (fuzzification) and decoding of the results (defuzzification) are steps that make possible to process images with fuzzy techniques. The main power of fuzzy image processing is in the middle step (modification of membership values). After the image data are transformed from gray-level plane to the membership plane (fuzzification), appropriate fuzzy techniques modify the membership values. It can be a fuzzy clustering, a fuzzy rule-based, fuzzy integration approach and so on [20].

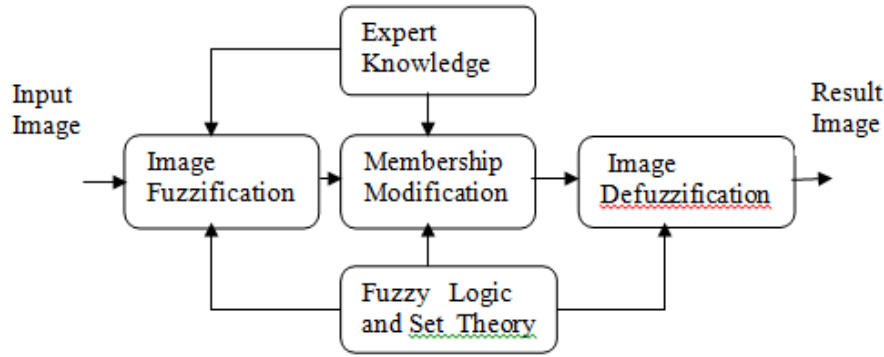

Figure 2.  The general structure of the fuzzy image processing

## 4.2. Steps involved in Fuzzy Logic Based Image Fusion

The fuzzy sets and fuzzy membership functions are required for system implementation was carried out considering that the input image and the output image obtained after defuzzification are both 8-bit quantized; this way, their gray levels are always between 0 and 255. The original image in the gray level plane is subjected to fuzzification and the modification of membership functions is carried out in the membership plane. The result is the output image obtained after the defuzzification process.

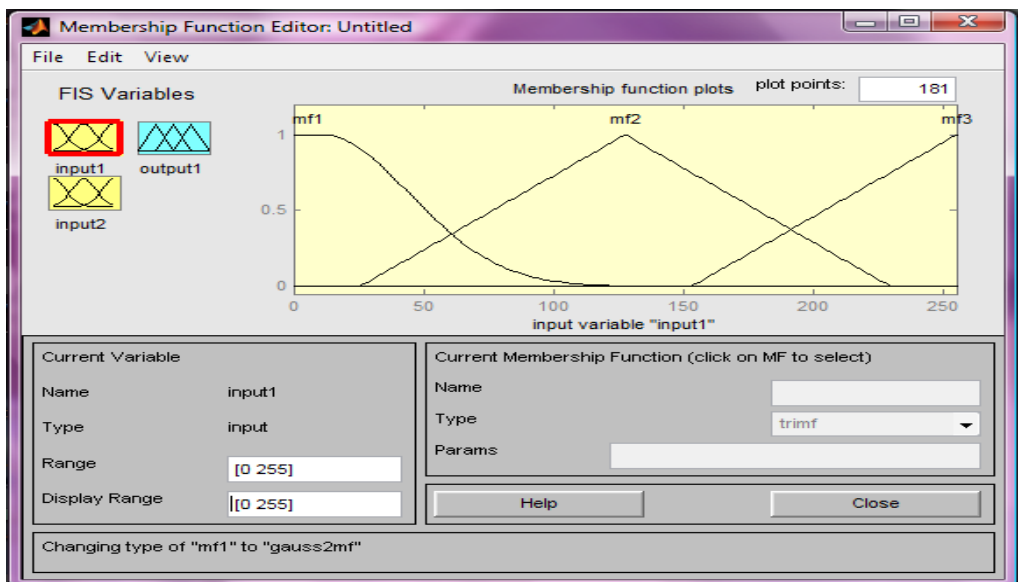

Figure  3. Applying membership functions for inputs

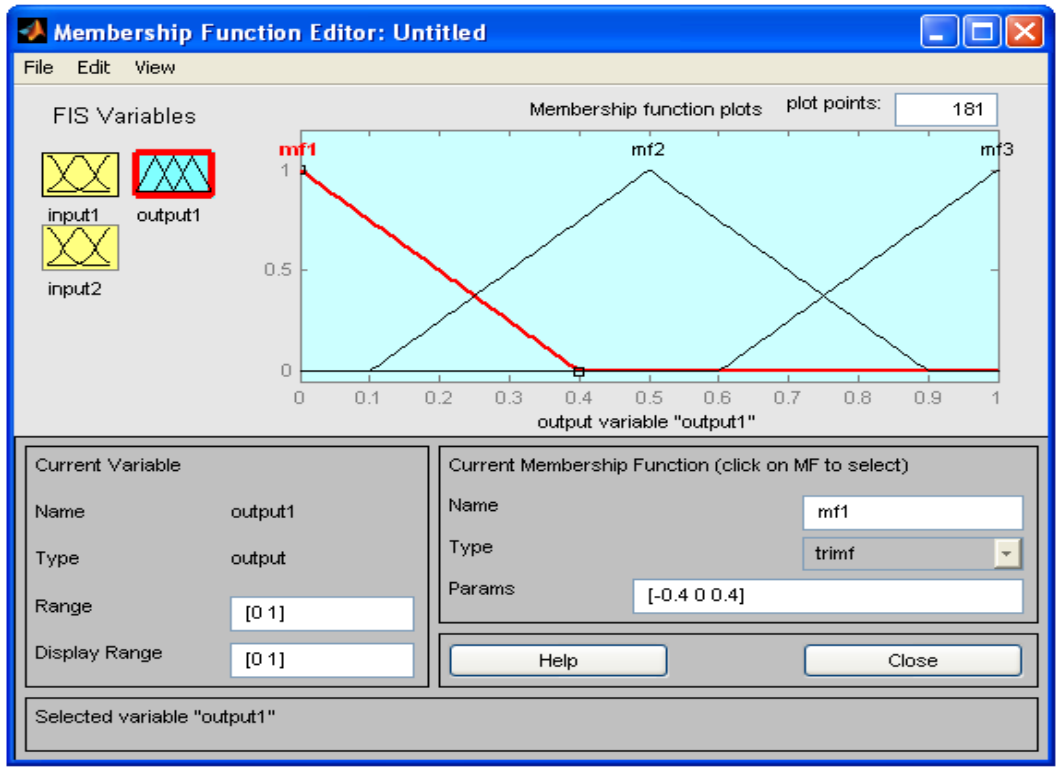

Figure 4. Applying membership functions for output

Rules considered in the fuzzy system

1. if (input1 is mf1) and (input2 is mf1) then (output1 is mf1)
2. if (input1 is mf2) and (input2 is mf1) then (output1 is mf2)
3. if (input1 is mf2) and (input2 is mf2) then (output1 is mf2)
4. if (input1 is mf3) or (input2 is mf2) then (output1 is mf3)
5. if (input1 is mf1) and (input2 is mf3) then (output1 is mf1)
6. if (input1 is mf3) or (input2 is mf3) then (output1 is mf2)

Algorithm steps for pixel level image fusion using Fuzzy Logic approach [21].

- Read first image in variable I1 and find its size (rows: rl, columns: c1).

- Read second image in variable I2 and find its size (rows:r2, columns: c2).

- Variables I1 and I2 are images in matrix form where each pixel value is in the range from 0-255. Use Gray Colormap.

- Compare rows and columns of both input images. If the two images are not of the same size, select the portion, which are of same size.

- Convert the images in column form which has C= rl*cl entries.

- Make a fis (Fuzzy) file, which has two input images.

- Decide number and type of membership functions for both the input images by tuning the membership functions. Input images in antecedent are resolved to a degree of membership ranging 0 to 255

- Make rules for input images, which resolve the two antecedents to a single number from 0 to 255.

- For num=l to C in steps of one, apply fuzzification using the rules developed above on the corresponding pixel values of the input images which gives a fuzzy set represented by a membership function and results in output image in column format

- Convert the column form to matrix form and display the fused image.

## 5. QUALITY EVALUATION MEASURES

Evaluation measures are used to evaluate the quality of the fused image. The fused images are evaluated, taking the following parameters into consideration.

### 5.1. Image Quality Index

IQI measures the similarity between two images (I1 & I2) and its value ranges from -1 to 1. IQI is equal to 1 if both images are identical. IQI measure is given by [22]

$$IQI = \frac{m_{ab} \cdot 2xy \cdot 2 m_a m_b}{m_a m_b \cdot x^2 + y^2 \cdot m_a^2 + m_b^2} \quad (4)$$

Where x and y denote the mean values of images I1 and I2 and $m_a^2$, $m_b^2$ and $m_{ab}$ denotes the variance of I1 , I2 and covariance of I1 and I2.

### 5.2. Mutual Information Measure

Mutual information measure furnishes the amount of information of one image in another. This gives the guidelines for selecting the best fusion method. Given two images M (i, j) and N (i, j) and MIM between them is defined as:

$$I_{MN} = \sum_{x,y} P_{MN}(x, y) \log \frac{P_{MN}(x, y)}{P_M(x) P_N(y)} \quad (5)$$

Where, $P_M(x)$ and $P_N(y)$ are the probability density functions in the individual images, and $P_{MN}(x, y)$ is joint probability density function

### 5.3. Fusion Factor

Given two images A and B, and their fused image F, the Fusion factor (FF) is illustrated as [23]

$$FF = I_{AF} + I_{BF} \tag{6}$$

Where $I_{AF}$ and $I_{BF}$ are the MIM values between input images and fused image. A higher value of FF indicates that fused image contains moderately good amount of information present in both the images. However, a high value of FF does not imply that the information from both images is symmetrically fused.

### 5.4. Fusion Symmetry

Fusion symmetry (FS) is an indication of the degree of symmetry in the information content from both the images.

$$FS = abs\left(\frac{I_{AF}}{I_{AF} + I_{BF}} - 0.5\right) \tag{7}$$

The quality of fusion technique depends on the degree of Fusion symmetry. Since FS is the symmetry factor, when the sensors are of good quality, FS should be as low as possible so that the fused image derives features from both input images. If any of the sensors is of low quality then it is better to maximize FS than minimizing it.

### 5.5. Fusion Index

This study proposes a parameter called Fusion index from the factors Fusion symmetry and Fusion factor. The fusion index (FI) is defined as

$$FI = I_{AF} / I_{BF} \tag{8}$$

Where $I_{AF}$ is the mutual information index between multispectral image and fused image and $I_{BF}$ is the mutual information index between panchromatic image and fused image. The quality of fusion technique depends on the degree of fusion index.

### 5.6. Root Mean Square Error

The root mean square error (RMSE) measures the amount of change per pixel due to the processing. The RMSE between a reference image R and the fused image F is given by

$$RMSE = \sqrt{\frac{1}{MN}\sum_{i=1}^{M}\sum_{j=1}^{N}(R(i,j) - F(i,j))} \tag{9}$$

### 5.7. Peak Signal to Noise Ratio

Peak signal to noise ratio (PSNR) can be calculated by using the formula

$$PSNR = 20\log_{10}\left[\frac{L^2}{MSE}\right] \tag{10}$$

Where MSE is the mean square error and L is the number of gray levels in the image.

### 5.8. Entropy

Entropy E, a scalar value representing the entropy of grayscale image. Entropy is a statistical measure of randomness that can be used to characterize the texture of the input image. Entropy is defined, Where p contains the histogram counts returned from imhist.

$$E = -sum(p * \log_2(p)) \qquad (11)$$

## 6. RESULTS AND DISCUSSIONS

There are many typical applications for image fusion. Modern spectral scanners gather up to several hundred of spectral bands which can be both visualized and processed individually, or which can be fused into a single image, depending on the image analysis task. In this section, input images are fused using fuzzy logic approach. Example 1, Panchromatic and Multispectral images of the Hyderabad city, AP, INDIA are acquired from the IRS 1D LISS III sensor at 05:40:44, Example 2 and Example 3 images are acquired from http://imagefusion.org [24].

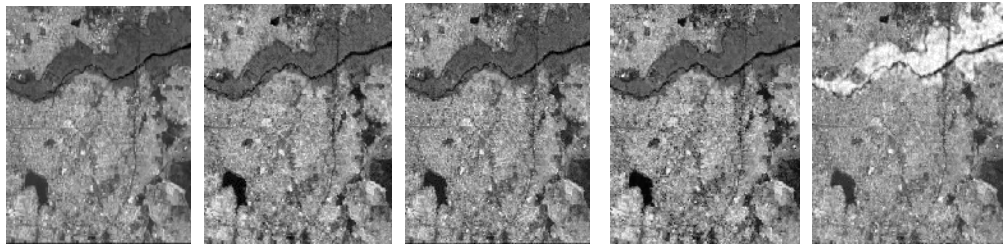

Example 1:    (a)          (b)          (c)          (d)          (e)

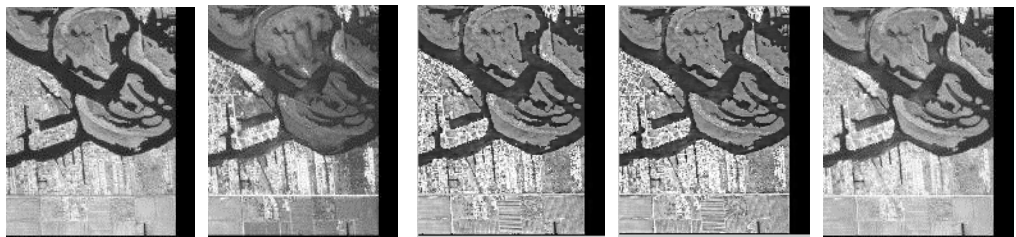

Example 2:    (f)          (g)          (h)          (i)          (j)

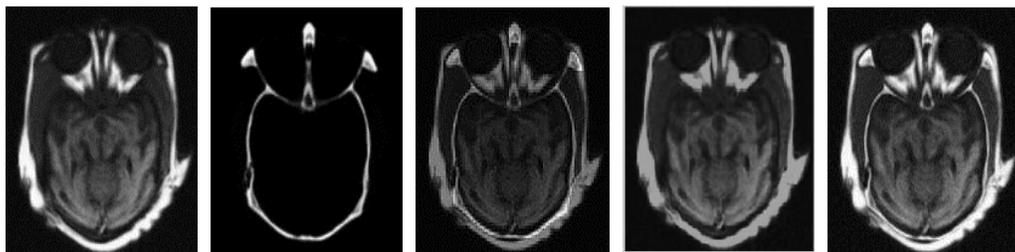

Example 3:    (k)          (l)          (m)          (n)          (o)

Figure 5.  Some example images (a), (b), (f), (g), (k) and (l): original input images;  (c), (h) and (m): fused images by wavelet transform, (d), (i) and (n):  fused images by weighted average DWT using GA and (e), (j) and (o): fused images by fuzzy logic respectively

Results obtained from above mentioned algorithms are given in Table 1.

Table 1.  The evaluation measures of image fusion based on Wavelet Transform, Weighted Average DWT using GA and Fuzzy Logic.

| Method | IQI | FF | FS | FI | MIM | RMSE | PSNR | Entropy |
|---|---|---|---|---|---|---|---|---|
| Wavelet Transform | | | | | | | | |
| (Example1) | 0.9473 | 3.8629 | 0.0429 | 1.1879 | 1.7656 | 63.5529 | 11.3425 | 7.3828 |
| (Example2) | 0.8650 | 3.8832 | 0.0118 | 0.9538 | 1.9875 | 19.1999 | 22.4648 | 7.2339 |
| (Example3) | 0.5579 | 2.6841 | 0.2731 | 3.4074 | 2.0751 | 39.5475 | 16.1884 | 5.9807 |
| Weighted Average DWT using GA | | | | | | | | |
| (Example1) | 0.9523 | 4.6519 | 0.03927 | 1.2841 | 2.1592 | 65.7253 | 11.7762 | 7.3571 |
| (Example2) | 0.9468 | 4.7282 | 0.01096 | 1.2179 | 1.0042 | 20.6849 | 21.8177 | 7.2418 |
| (Example3) | 0.7374 | 3.2841 | 0.3647 | 4.6382 | 3.2743 | 35.726 | 17.0711 | 6.4248 |
| Fuzzy Logic | | | | | | | | |
| (Example1) | 0.9689 | 5.5687 | 0.2752 | 3.4475 | 4.3166 | 52.5301 | 13.7226 | 7.3445 |
| (Example2) | 0.9955 | 8.8407 | 0.0598 | 1.2719 | 3.8914 | 17.8385 | 23.1036 | 7.2577 |
| (Example3) | 0.9896 | 4.7589 | 0.4023 | 9.2320 | 4.2938 | 25.4703 | 20.0101 | 6.7300 |

The proposed algorithm has been implemented using Matlab 7.0. Proposed fuzzy based image fusion approach can be scalable and expandable for great many situations like remote sensing and medical imaging in which membership functions and rules have to define precisely used in fuzzy inference system. In order to evaluate the fusion results obtained from different methods and compare the methods, the assessment measures are employed**.** The value of each quality assessment parameters of all mentioned fusion approaches are depicted in Table 1.Our experimental results show that proposed fuzzy logic based image fusion approach provides better performance and quality on compared to conventional wavelet transform (WT) and weighted average discrete wavelet transform based image fusion using genetic algorithm (GA ) techniques. Due to efficiency of the proposed method the image quality index (IQI), the similarity between reference and fused image (0.9689, 0.9955, 0.9896) are higher for three cases in fuzzy based fusion technique compared to IQI values (0.9473, 0.8650, 0.5579) from wavelet transform and IQI values (0.9523, 0.9468, 0.7374) from weighted average DWT based image fusion using GA. The higher values (5.5687, 8.8407, 4.7589) of fusion factor (FF) obtained from the fuzzy based fusion approach indicates that fused image contains moderately good amount of information present in both the images compared to FF values (3.8629, 3.8832, 2.6841) obtained from wavelet transform and FF values (4.6519, 4.7282, 3.2841) obtained from weighted average DWT based image fusion using GA. The amount of information of one image in another, mutual information measure (MIM) is also significantly better which shows that proposed fuzzy based fusion method preserves more information compared to  conventional wavelet transform  and weighted average discrete wavelet transform based image fusion using genetic algorithm. The other evaluation measures like root mean square error (RMSE) with lower and peak signal to noise ratio (PSNR) with higher values are also comparatively better for fuzzy based fusion approach. Finally entropy, the amount of information that can be used to characterize the input image also better for images obtained from fuzzy based image fusion technique. So it is concluded that results obtained from the implementation of fuzzy logic based image fusion approach performs better than wavelet transform based image fusion and weighted average DWT based image fusion using GA through typical assessment parameters.

## 7. CONCLUSIONS

There are a large number of applications in remote sensing that require images with both spatial and spectral resolution. In this paper, the potentials of pixel level image fusion using fuzzy logic approach has been explored along with quality assessment evaluation measures. Fused images are primarily used to human observers for viewing or interpretation and to be further processed by a computer using different image processing techniques**.** All the results obtained and discussed by this method are same scene. The experimental results clearly show that the introduction of the proposed image fusion using fuzzy logic gives a considerable improvement on the quality of the fusion system. It is hoped that the technique can be further extended to all types of images, for fusion of multiple sensor images and to integrate valid evaluation measures of image fusion using neuro fuzzy logic. Future work also includes the iterative fuzzy logic and neuro fuzzy logic, which efficiently gives good results.

## ACKNOWLEDGEMENTS

This work was partially supported by the All India Council for Technical Education, New Delhi, India under Research Promotion Scheme, Grant No. 8023/RID/RPS-80/2010-11.

**Authors**

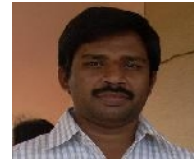

**Srinivasa Rao Dammavalam**  received his  M. Tech in Information Technology from JNTU, Hyderabad and pursuing his Ph.D from JNTU, Kakinada. He worked as a research fellow at **AVIRES Lab, University of Udine, Italy** for two years. Currently he is working as a Senior Assistant Professor in the Department of Information Technology in VNRVJIET, Hyderabad. He was the recipient of the **AICTE** grant for his proposal under **Research Promotion Scheme** over a period of two years.

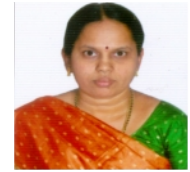

*Dr. M.Seetha.* received her Ph.D., in Computer Science  and Engineering in the area of image processing  from J.N.T.Universty, Hyderabad, India. She is currently working as a Professor in the Dept. of CSE, GNITS, Hyderabad. She was the recipient of the **AICTE Career Award for Young Teachers (CAYT)** in FEB, 2009, and received the grant upto **10.5 lakhs** over a period of three years by AICTE, INDIA. She was a reviewer for various International Journals/Conferences. Her research interests include image processing, neural networks, computer networks, artificial intelligence and data mining.

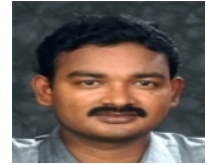

**MHM Krishna Prasad** received his Ph.D., in Specific Data Mining and in general Computer Science & Engineering from J.N.T.Universty, Hyderabad, India. He is currently holding the position of Associate Professor & Head of Department of Information Technology, University College of Engineering – JNTUK, Vizianagaram campus. His research interests include Data Mining, Network Security and Image Processing.